\documentclass[10pt,twocolumn,letterpaper]{article}

\usepackage{algorithm}
\usepackage{algorithmic}
\usepackage{adjustbox} 
\usepackage{booktabs}
\usepackage{diagbox}
\usepackage{multirow}
\usepackage{amsmath}
\usepackage{amsfonts}
\usepackage{amssymb}
\usepackage{makecell}
\usepackage{subfigure}
\usepackage{colortbl}
\usepackage{xcolor}
\usepackage{tcolorbox}
\usepackage[switch]{lineno}

\usepackage[final]{cvpr}      

\definecolor{cvprblue}{rgb}{0.21,0.49,0.74}
\usepackage[pagebackref,breaklinks,colorlinks,citecolor=cvprblue]{hyperref}

\def\paperID{******} 
\def\confName{CVPR}
\def\confYear{2024}

\title{LiDAR-LLM: Exploring the Potential of Large Language Models\\for 3D LiDAR Understanding}

\author{
Senqiao Yang\textsuperscript{\rm 1*},
Jiaming Liu\textsuperscript{\rm 1,2}\thanks{Equal contribution: \{yangsenqiao.ai,liujiaming.pku\}@gmail.com}, 
Ray Zhang \textsuperscript{\rm 3*}\thanks{Project leader},
Mingjie Pan\textsuperscript{\rm 1*},
Zoey Guo \textsuperscript{\rm 3},
Xiaoqi Li \textsuperscript{\rm 1},\\
Zehui Chen \textsuperscript{\rm 1},
Peng Gao \textsuperscript{\rm 3},
Yandong Guo\textsuperscript{\rm 2},
Shanghang Zhang\textsuperscript{\rm 1}\thanks{Corresponding author: shanghang@pku.edu.cn}
\vspace{0.1cm}\\
\textsuperscript{\rm 1}National Key Laboratory for Multimedia Information Processing, School of Computer Science,\\ Peking University
\textsuperscript{\rm 2}AI2Robotics  \textsuperscript{\rm 3} Shanghai Artificial Intelligence Laboratory\\
}
\vspace{-0.1cm}
\begin{document}
\maketitle
\begin{abstract}

Recently, Large Language Models (LLMs) and Multimodal Large Language Models (MLLMs) have shown promise in instruction following and 2D image understanding. While these models are powerful, they have not yet been developed to comprehend the more challenging 3D physical scenes, especially when it comes to the sparse outdoor LiDAR data. In this paper, we introduce LiDAR-LLM, which takes raw LiDAR data as input and harnesses the remarkable reasoning capabilities of LLMs to gain a comprehensive understanding of outdoor 3D scenes. The central insight of our LiDAR-LLM is the reformulation of 3D outdoor scene cognition as a language modeling problem, encompassing tasks such as 3D captioning, 3D grounding, 3D question answering, etc. Specifically, due to the scarcity of 3D LiDAR-text pairing data, we introduce a three-stage training strategy and generate relevant datasets, progressively aligning the 3D modality with the language embedding space of LLM. Furthermore, we design a View-Aware Transformer (VAT) to connect the 3D encoder with the LLM, which effectively bridges the modality gap and enhances the LLM's spatial orientation comprehension of visual features. Our experiments show that LiDAR-LLM possesses favorable capabilities to comprehend various instructions regarding 3D scenes and engage in complex spatial reasoning. LiDAR-LLM attains a 40.9 BLEU-1 on the 3D captioning task and achieves a 63.1\% classification accuracy and a 14.3\% BEV mIoU on the 3D grounding task. Web page: \href{https://sites.google.com/view/lidar-llm}{https://sites.google.com/view/lidar-llm}


\end{abstract}    
\section{Introduction}
\label{sec:intro}

\begin{figure}[htp]
\includegraphics[width=0.48\textwidth]{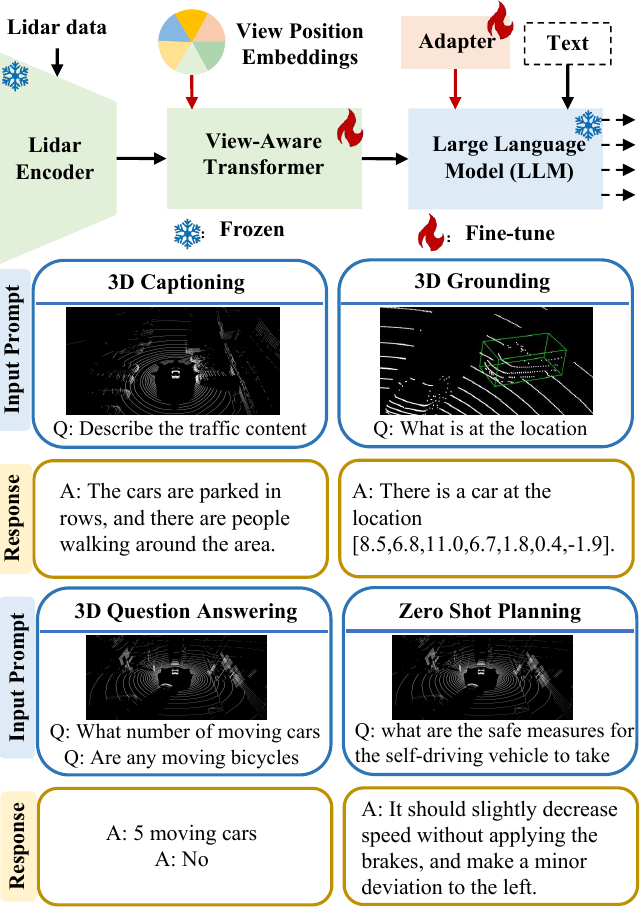}
\vspace{-0.4cm}
\centering
\caption{
\textbf{Characteristics of LiDAR-LLM.}
Our proposed LiDAR-LLM takes 3D LiDAR data as input and aligns the 3D modality with the language embedding space, leveraging the exceptional reasoning capabilities of LLMs to understand outdoor 3D scenes. To enhance the spatial orientation representation of LiDAR features, we propose a View-Aware Transformer between the LiDAR Encoder and LLMs.
Simultaneously, the bottom part showcases examples derived from our generated or employed LiDAR-text data, covering a spectrum of 3D-related tasks.
}
\label{fig: intro}
\vspace{-0.4cm}
\end{figure}

Recently, large language models (LLMs) \cite{touvron2023llama, openai2023gpt4, brown2020language} have demonstrated significant capabilities in complex reasoning and robust conversational abilities in the field of natural language processing. Building upon LLMs, Multimodal Large Language Models (MLLMs), such as BLIP-2 \cite{li2023blip2} and Flamingo \cite{Alayrac2022Flamingo}, have been introduced. These models take in more modality (e.g., 2D images) as input, enabling LLMs to discuss and comprehend the visual scene. 
Despite MLLMs excelling at processing 2D image content, their comprehension of the more challenging 3D real-world scenes remains an open question. 
Understanding 3D scenes holds importance for various applications, including autonomous driving \cite{arnold2019survey, chi2023bev, pan2023renderocc, chen2022bevdistill} and robotics \cite{dhamo2021graph, yao20183d, li2023imagemanip, wang2023find}, due to the wealth of spatial information in 3D data.


Existing 3D understanding methods \cite{yang2021sat, jiao2022more, parelli2023clip, azuma2022scanqa, ma2022sqa3d} often fail to demonstrate sufficient generalization capabilities when faced with unseen scenarios. They are limited in expressing specific downstream tasks in a manner comprehensible to humans, such as generating scene captioning and question answering.
Therefore, recent works \cite{wang2023chat, hong20233d, guo2023point} take indoor 3D point clouds as input and leverage the powerful capabilities of LLMs to analyze them, aligning the 3D characteristics with the textual features of LLMs. However, they still encounter challenges when dealing with 3D outdoor LiDAR data, primarily due to its sparsity and complex geometric relationships, which pose a difficult multimodal alignment and reasoning.

In this paper, as shown in Figure \ref{fig: intro}, we introduce LiDAR-LLM, a novel approach that harnesses the reasoning capabilities of LLMs to comprehensively understand outdoor 3D scenes.
The LiDAR-LLM architecture comprises a 3D LiDAR encoder, an intermediate alignment transformer, and an
LLM, \textit{e.g.}, LLaMA~\cite{touvron2023llama}.
The key insight of LiDAR-LLM lies in redefining the problem of 3D scene cognition through interpretative language modeling. However, the introduction of LLMs for perceiving outdoor 3D scenes faces two challenges: (1) In contrast to the abundant availability of image-text paired data \cite{sharma2018conceptual, schuhmann2022laion, changpinyo2021conceptual}, 3D LiDAR-text paired data is exceedingly rare, and readily accessible multimodal models (e.g., CLIP \cite{radford2021learning}) are lacking. (2) 3D LiDAR data encompasses a variety of objects and intricate geometric relationships among them. Take outdoor autonomous driving, for example, where the ego vehicle is surrounded by a diverse array of moving and stationary objects, which both occlude and influence each other.

To tackle these challenges, for LiDAR-LLM, we introduce a three-stage training strategy and generate relevant datasets, gradually transferring 3D representations into the text feature space and unleashing LLMs' reasoning capabilities for 3D scenes. Specifically, in the first stage, we employ MLLMs \cite{zhang2023llama, li2023blip2} and GPT4 \cite{openai2023gpt4} for communication between multi-view images and language within the nuScenes dataset \cite{nuscenes}, where each scene is accompanied by paired 3D LiDAR data.
In this way, we generate a dataset of 420K liDAR-text pairs and \textbf{cross-modal align} the 3D LiDAR features with the word embeddings of LLMs. 
During the second stage, as the \textbf{perception} forms the foundation of 3D scene understanding, we incorporate the 3D bounding boxes into the question-answer text and generate a 280K LiDAR grounding dataset. This enhances LiDAR-LLM's sensitivity to object locations and relationships. 
In the final stage, we perform efficient fine-tuning of our model on \textbf{high-level instruction} datasets \cite{qian2023nuscenes, drivelm2023}, comprehensively expanding its capabilities for 3D downstream tasks.
To more effectively bridge the modality gap between 3D LiDAR and text, we design a View-Aware Transformer (VAT) that connects the 3D LiDAR encoder with the LLM, injecting six view position embeddings into the 3D feature. 
Combined with the three-stage training strategy, VAT enhances the LLM's comprehension of the spatial orientation of visual features.
In summary, our contributions are as follows:

\begin{itemize}
\item 
We propose LiDAR-LLM, which takes 3D LiDAR data and language prompts as input, harnessing the reasoning capabilities of LLMs to understand outdoor 3D scenes. LiDAR-LLM can perform tasks such as 3D captioning, 3D grounding, 3D question answering, and more.
\item
We introduce a three-stage training strategy for gradually transferring 3D representations into the text feature space, which involves cross-modal alignment, perception, and high-level instruction. In parallel, we collect a set of LiDAR-text paired datasets, including 420K 3D captioning and 280K 3D grounding data, which will be released.

\item 
We specially design a View-Aware Transformer (VAT) that connects the 3D LiDAR encoder with the LLM, bridging the modality gap between 3D LiDAR and text, and enhancing the LLM's comprehension of the spatial orientation of visual features.

\item 
In our proposed LiDAR-text datasets, LiDAR-LLM exhibits superior performance, achieving a 40.9 BLEU-1 score on the 3D captioning dataset, and securing a classification accuracy of 63.1\% and a BEV mIoU of 14.3\% on the 3D grounding dataset.

\end{itemize}

\begin{figure*}[t]
\centering
\includegraphics[width=\linewidth]{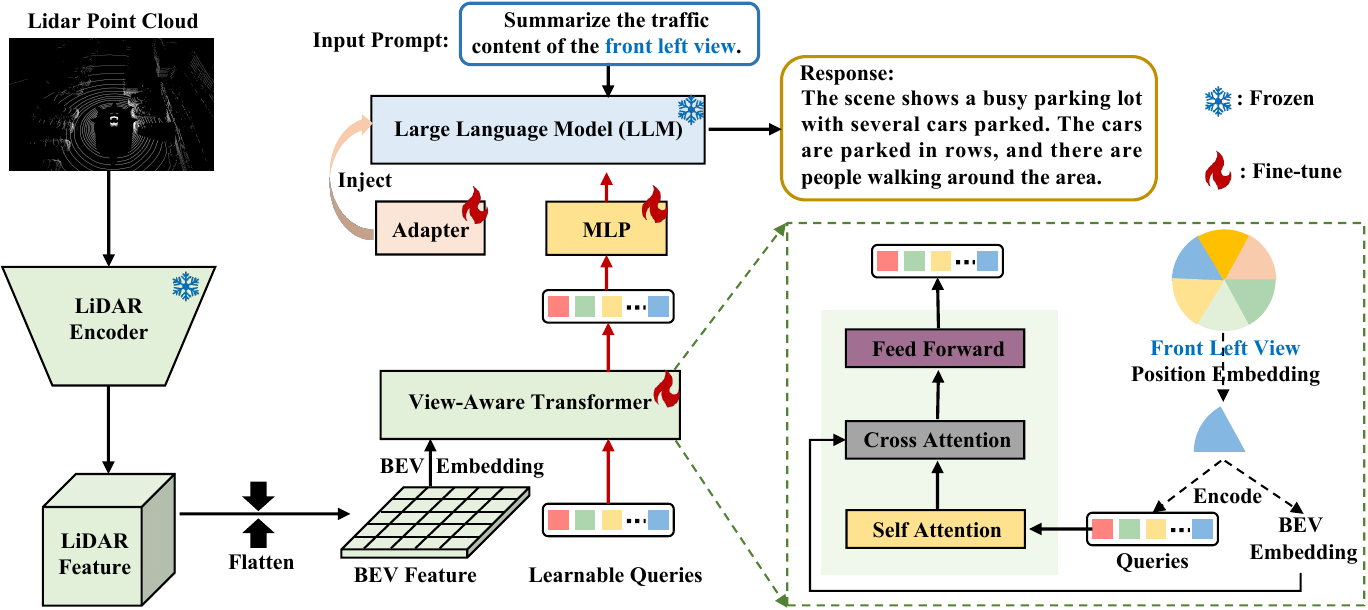}
\vspace{-0.4cm}
\caption{
\textbf{Overview of our LiDAR-LLM framework.} 
The initial column showcases our 3D feature extractor, which processes the LiDAR point cloud input to derive a 3D voxel feature. Subsequently, the feature is flattened along the z-axis to produce the bird’s-eye view (BEV) feature. The View-Aware Transformer (VAT) accepts BEV embedding and learnable queries as input, with the output queries serving as soft prompt input to the frozen LLM. In the VAT, we introduce six view position embeddings into the BEV feature along with corresponding queries to enhance the capability of spatial orientation representation. This framework aligns the LiDAR modality with the language embedding space, enabling us to leverage the LLM for a comprehensive understanding of outdoor 3D scenes. }
\label{fig: framework} 
\vspace{-0.1cm}
\end{figure*}

\section{Related Work}
\subsection{Multi-modal Large Language Models}
Extensive language models, such as LLaMA~\cite{touvron2023llama} and GPT-3~\cite{floridi2020gpt}, demonstrate proficiency in managing a variety of language tasks, leveraging their powerful reasoning and generalization abilities.
Building upon these achievements, 2D Multi-modal Large Language Models (2D MLLMs)~\cite{zhang2023llama,Alayrac2022Flamingo,li2023blip2} are introduced to bridge RGB visual images and text. These models leverage the capabilities of Large Language Models (LLMs)~\cite{touvron2023llama} and, by conditioning on 2D inputs, aim to address 2D downstream tasks, such as visual question answering~\cite{antol2015vqa} and captioning~\cite{anderson2018bottom}. The representative model, BLIP~\cite{li2023blip2}, pre-trains a multimodal mixture of encoder-decoder models using a dataset bootstrapped from large-scale noisy image-text pairs. It injects diverse synthetic captions and removes noisy captions to achieve unified vision-language understanding and generation. 
Meanwhile, VisionLLM \cite{wang2023visionllm} aligns vision-centric tasks with language tasks, allowing for flexible definition and management through language instructions. 
Furthermore, the introduction of 3D Multi-modal Large Language Models (3D MLLMs)~\cite{guo2023point,hong20233d, wang2023chat, xu2023pointllm} aims to broaden the scope of knowledge, reasoning, and conversational capabilities obtained from LLMs to encompass the 3D modality. For instance, several projects leverage GPT-3 \cite{floridi2020gpt} or LLaMA \cite{touvron2023llama} to improve the language-based comprehension of 3D spatial geometry, as demonstrated in works like PointCLIP V2 \cite{zhu2022pointclip} and ViewRefer \cite{guo2023viewrefer}. They focus on the 3D point cloud with a single object or an indoor scene.

In contrast to these approaches, we are the first to exploit the reasoning capabilities of LLMs for understanding outdoor 3D scenes and completing tasks such as captioning, 3D grounding, and 3D question answering. The unique challenges posed by 3D LiDAR point cloud data, including the lack of LiDAR-text paired data and encompassing a variety of objects and relationships, present difficulties in multimodal alignment and reasoning. 

\subsection{3D-Language Tasks}
The combination of 3D point clouds and natural language holds diverse applications and has recently attracted growing attention \cite{chen2021scan2cap,achlioptas2020referit3d,feng2021free,hong20223d,huang2021text,chen2020scanrefer}. Specifically, 3D captioning~\cite{chen2023unit3d, chen2021scan2cap} is required to describe a particular object within a 3D scene. 3D visual grounding~\cite{yang2021sat,chen2020scanrefer} focuses on generating the location of the object that the text expression refers to. Meanwhile, in the context of 3D visual question answering~\cite{azuma2022scanqa}, the model needs to answer language questions given the visual content of the 3D scene.

However, 3D approaches for the aforementioned tasks are designed to address individual task-specific challenges without exploring their commonalities and providing a unified solution. Moreover, these methods are tailored for indoor point cloud tasks and may not directly transfer to outdoor LiDAR since LiDAR is much sparser and more diverse in geometric relationships. To reconcile this, we propose LiDAR-LLM, a LiDAR-oriented approach, to uniformly perform 3D tasks of outdoor scenes.

\section{Method}
\label{sec:method}

\subsection{Overview}
The overall framework of LiDAR-LLM is presented in Fig.~\ref{fig: framework}. The core concept involves transforming the highly sparse and intricate geometric LiDAR data into the representation space understandable by Large-Language Models (LLMs). This transformation is facilitated by our proposed View-Aware Transformer (VAT), which incorporates view position embeddings to enhance the spatial orientation understanding of the LLM. Thus, it enables a comprehensive interpretation of intricate details in the outdoor 3D scene.

However, the integration of LLMs to comprehend outdoor 3D scenes faces two challenges: (1) Unlike the abundance of available image-text paired data, 3D LiDAR-text paired data is exceptionally scarce; and (2) 3D LiDAR data involves diverse objects and intricate geometric relationships among them.
Therefore, we implement a three-stage training strategy and generate LiDAR-text paired training data to collaboratively align 3D representations with the feature space of LLMs.
Through this process, LiDAR-LLM undergoes diverse tasks across modalities and handles complex cross-modal scenarios at both the scene and instance levels. It unleashes the LLMs' common-sense reasoning and localization capabilities on 3D LiDAR data.

\subsection{Model Architecture}
Given a LiDAR input $L\in \mathbb{R}^{n\times 3}$, where 
$n$ is the number of points, VoxelNet \cite{zhou2018voxelnet} is employed to extract its 3D voxel feature. Subsequently, considering the computational cost, we flatten the feature along the z-axis to generate the bird’s-eye view (BEV) feature. Simultaneously, for the text input $T$ with a maximum of $m$ characters, LLaMA \cite{touvron2023llama} is utilized to extract text features.
With the BEV feature $\mathcal{F}_{v}\in \mathbb{R}^{c\times h \times w}$ along with the text feature $\mathcal{F}_{t}\in \mathbb{R}^{m\times d}$ (where $d$ is the dimension of the feature), our objective is to project these LiDAR BEV features into the word embedding space of a pre-trained LLaMA through our proposed View-Aware Transformer (VAT). This alignment is crucial for conducting multi-modal understanding and generating accurate answers in 3D downstream tasks. During training, we only fine-tune the injected adapters \cite{hu2021lora} in the LLaMA and VAT module while freezing the major parameters. This aims to preserve the powerful feature extraction and reasoning ability of existing modules and further equip the model with capabilities in understanding 3D LiDAR scenes.

\textbf{VAT design.}
In the right part of Fig. \ref{fig: framework}, the input to the VAT includes a set of $K$ learnable query embeddings, with $K$ set to 576 for convenient projection into the word embedding space of the LLM. These queries interact with the BEV feature through a cross-attention mechanism. The VAT produces an output comprising $K$ encoded visual vectors, one for each query embedding. These vectors then undergo processing through a multi-layer perceptron (MLP) and are subsequently fed into the frozen LLM.
However, outdoor LiDAR data, such as nuScenes \cite{nuscenes}, demands a comprehensive understanding of the orientation relationships between diverse objects and the ego car. And it encompasses intricate relationships among objects. Hence, we introduce view position embedding for the BEV feature, with the aim of promoting the model's capacity to learn orientation and geometric relationships.
Specifically, we first construct the view position embedding $\mathcal{V}_{p}\in \mathbb{R}^{c\times 6}$ with zero initial parameters. Then, we split the BEV feature according to six views, including front, front right, front left, back, back right, and back left view. During training, when dealing with a question related to a specific view, we inject the corresponding position embedding into both the BEV feature and queries. For instance, when training a caption sample related to the front left view, we only inject the front left position embedding $\mathcal{V}_{p}\in \mathbb{R}^{c\times 1}$ into the front left view portion of the BEV feature and queries. 
If the training sample involves a question regarding the entire panoramic scene, we inject all six view position embeddings during training.



\subsection{Three-stage training strategy}
\label{sec:stage}
In this section, we demonstrate how we empower LLMs with the capabilities to comprehend 3D LiDAR data and uniformly complete extensive 3D tasks. We introduce a three-stage training strategy and generate relevant datasets, gradually transferring 3D representations into the text feature space. Three stages contain cross-modal alignment, perception, and high-level instruction.
 
\textbf{Cross-Modal Alignment (3D Captioning):}
To effectively address abundant 3D downstream tasks, the model requires a thorough understanding of the LiDAR scene.
Scene captioning is a logical approach to enable the model to capture essential information and details in the LiDAR data by integrating the entire 3D scene into LLMs. 

However, the absence of direct LiDAR and text description pairs for caption training motivates us to leverage existing multi-view images aligned with LiDAR data in nuScenes \cite{nuscenes} for creating text descriptions.
Employing powerful off-the-shelf 2D Multi-Modal LLMs (MLLMs) \cite{li2023blip2, zhang2023llama}, we generate captions for each view, creating textual descriptions corresponding to the LiDAR scene.
Nevertheless, the captions for LiDAR data and 2D multi-views are not perfectly aligned, as 2D MLLM may provide descriptions related to weather or colors for 2D images, which are not applicable to LiDAR data.
To address this inconsistency, we further employ GPT-4~\cite{openai2023gpt4} to filter out captions that are more relevant and suitable for LiDAR data.

With the collected LiDAR-caption pairs, our goal is to enable LLaMA to generate descriptive text conditioned on LiDAR input. 
We observe that textual captions for LiDAR data tend to be excessively detailed and lengthy due to their intricate geometric structures.  
Jointly learning overall captions could lead to entanglement in LLM reasoning.
To mitigate this, we initially train the model to caption a single view to reduce complexity.
The output caption is supervised by the ground-truth answer of the corresponding view using cross-entropy loss.
After enabling the model to acquire captioning skills for individual views, the subsequent step involves instructing the model to understand the entire panoramic scene and generate a global description.
By doing so, we align the 3D feature representation to the text feature space of LLM, enabling the model to comprehend the context in the LiDAR data.

\textbf{Perception:}
After equipping the model with a global scene understanding, this stage focuses on endowing the model with instance-level perception abilities, as they form the foundation for high-level instructional tasks such as planning.
To achieve this, we employ an object-centric learning strategy, ensuring the model is cognizant of various object details such as quantity, localization, and spatial relations. The model learns the alignment between the representation of individual 3D objects and the corresponding text embedding of the LLM associated with the objects.

Two tasks, visual grounding, and grounded captioning, are designed for this purpose. Objects are first represented as a sequence of discrete tokens, where each object's label and bounding box are extracted.
Given a 3D object with its annotations, the category name and locations are encoded into a word embedding using the tokenizer of the pre-trained LLM. 
Unlike the previous indoor 3D MLLM \cite{wang2023chat}, there is no need to extract each object from the point cloud individually; instead, we achieve object perception across the entire 3D scene.
For visual grounding, the model learns to generate location tokens specifying the region position $(x_{1},y_{1},z_{1},x_{2},y_{2},z_{2}, \theta)$ based on the LiDAR input and instruction, where $\theta$ is the box angle. 
The task of Grounded Captioning is positioned as the inverse counterpart to visual grounding. The model is trained to generate descriptive text by leveraging the input LiDAR data and text with location information.
Outputs of both tasks are subject to supervision through cross-entropy loss. The formulations of the instructions are depicted in Fig. \ref{fig: vis}. This alignment process aims to align the 3D visual object embedding with the text embedding space, unlocking LLM's 3D perception ability.

\textbf{High-level Instruction:}
In this stage, having comprehensively understood the LiDAR scene and equipped the model with basic 3D perception capabilities, we leverage a high-level instruction dataset (e.g., nuScenes-QA \cite{qian2023nuscenes}) to further enhance the model's reasoning skills in the 3D space. 
Through the fine-tuning of LiDAR-LLM using this dataset, we not only enhance its proficiency in comprehending a diverse array of instructions but also empower it to generate responses that are both creative and contextually appropriate.
Moreover, this refinement process equips LiDAR-LLM with the capability to engage in intricate spatial reasoning and integrate external knowledge into its generated responses. These tasks are also supervised through cross-entropy loss, ensuring the effective alignment of the model's output with the desired high-level instructions. 
Meanwhile, we also explore the autonomous driving planning capabilities of LiDAR-LLM on the nuScenes dataset \cite{nuscenes}. Instead of generating any planning QA data, we directly utilize our trained model to infer questions related to planning. We found that, through our proposed three-stage training strategy, LiDAR-LLM develops preliminary planning capabilities, as illustrated in Fig. \ref{fig: vis}. 
The results also demonstrate that our training pipeline can stimulate the model's reasoning ability in 3D LiDAR data.

\subsection{Training and Task Inference.}

LiDAR-LLM undergoes joint fine-tuning with a variety of tasks and datasets, equipping it with a versatile skill set to adeptly handle diverse tasks within complex cross-modal scenarios. In the fine-tuning phase, we perform fine-tuning on a dataset consisting of 700K LiDAR-text pairs generated by us and 460K publicly available datasets \cite{qian2023nuscenes}. Throughout the training process, the mentioned tasks are systematically trained in a step-by-step manner.
During inference, our input still consists of LiDAR and question text. We have the flexibility to infer each question individually or consecutively infer multiple questions.
\section{Experiment}
In this section, we conduct extensive experiments on nuScenes \cite{nuscenes}, nuScenes-QA \cite{qian2023nuscenes}, and our generated datasets. We first introduce the selected baseline and evaluation metrics (\S \ref{sec:baselines}), as well as the implementation details (\S \ref{sec:exp_details}). Our main experiments evaluate the model with three essential capabilities: 3D captioning (\S \ref{sec:exp1}), 3D grounding (\S \ref{sec:exp2}), and holistic high-level instruction following (\S \ref{sec:exp3}). Finally, we present detailed ablation studies (\S \ref{sec:abla}) and qualitative analysis (\S \ref{sec:vis}) to provide a deeper understanding of our approach. Additional quantitative and qualitative analyses are available in Appendices A and B, respectively.


\begin{table*}[t]
\begin{center}
\centering
    \setlength\tabcolsep{0.25cm}
    \small
    \begin{tabular}{c|c|cccccc}
    \toprule
    Tasks& Models   & BLEU-1& BLEU-2& BLEU-3& BLEU-4 & Bert Score \\
\hline
\multirow{5}{*}{3D Captioning}& Mini-GPT4~\cite{zhu2023minigpt} & 14.97   & 6.76   & 3.74 & 2.63 &84.38\\
 & LLaVA1.5~\cite{liu2023improvedllava}  & 19.92  &12.10 &8.57&5.37  &85.01 \\
 & Instruct-BLIP~\cite{li2023blip2}  & 18.67 &13.38 &7.41&5.20  &85.89 \\
 & LLaMA-AdapterV2~\cite{zhang2023llama}  & 30.17  &17.34 &10.40&7.45  &86.45 \\
 
 & Ours & 40.98& 29.96 & 23.43 & 19.26   &91.32\\
    \bottomrule
    \end{tabular}
\end{center}
\vspace{-0.3cm}
\caption{\label{tab: cap} 
    Experimental Results on nu-Caption dataset. Our model outperforms all baseline models for all evaluation metrics.}
\end{table*}

\subsection{Baselines \& Evaluation Metrics}
\label{sec:baselines}
\textbf{Baselines.} 
To the best of our knowledge, we are the first Multi-modal LLM (MLLM) to utilize LiDAR data with textual instructions as input and implement a series of outdoor 3D tasks. Since there are no 3D MLLM that can directly process LiDAR data, we project the depth information from LiDAR onto the 2D plane and employ current state-of-the-art (SOTA) 2D MLLM methods, MiniGPT-4 \cite{zhu2023minigpt} and LLaMA-Adapter V2 \cite{zhang2023llama}, as our competitive counterparts.\\
\textbf{Evaluation Metrics.} To assess the quality of language generation in the 3D Captioning Task, we adopt both BLEU \cite{papineni2002bleu} and BERT Score \cite{zhang2019bertscore} to comprehensively gauge the response quality from the model. 
In the 3D Grounding Task, we evaluate the grounding ability of our proposed models through a combination of classification Top-1 accuracy and BEV mean Intersection over Union (mIoU). For NuScene-QA (high-level instruction), we assess model performance using Top-1 accuracy, in line with common practices in Visual Question Answering (VQA) research \cite{antol2015vqa, azuma2022scanqa}. We also conduct separate evaluations for different question types.

\subsection{Implementation Details}
\label{sec:exp_details}

Our LiDAR-LLM mainly comprises three components: LiDAR feature extraction backbone, View-Aware Transformers (VAT), and the Large Language Model (LLM). 
For the LiDAR feature extraction, we employ the standard pre-trained 3D detector, CenterPoint-Voxel \cite{zhou2018voxelnet} following its default settings. The point cloud range is [-54.0m, 54.0m, -5.0m, 54.0m, 54.0m, 3.0m], and the BEV grid size is [0.6m, 0.6m].
For the VAT, we set the token number of learnable queries to 576, and the dimension of the token is 768. 
In terms of the LLM, we employ LLaMA-7B \cite{touvron2023llama} considering both efficiency and efficacy. 
Throughout the three-stage training phase, we utilize the Adam optimizer $(\beta_{1}, \beta_{2}) = (0.9, 0.999)$ with an initial learning rate of 1e-4, halving it every 2 epochs. And we fine-tuning the VAT and adapters in LLaMA2 for 6 epochs.
All experiments are conducted on NVIDIA Tesla A100 GPUs.

\subsection{3D Captioning}
\label{sec:exp1}

\textbf{Dataset Construction.} Due to the absence of a captioning dataset tailored for LiDAR data, we integrate GPT-4 \cite{openai2023gpt4} and 2D MLLMs \cite{zhang2023llama, li2023blip2} to construct a large-scale 3D captioning dataset on nuScenes (named nu-Caption), which consists of 420K high-quality LiDAR-text pairs. 
In nu-Caption, we employ 348K LiDAR-text pairs for the training set and 72K pairs for the validation set.
The caption question covers three aspects, progressing from simple to difficult: 1) the general description of current scenes or traffic conditions, 2) the detailed description of objects and their relationships, 3) and the recognition of potential risks on the road.
Concrete examples are depicted in Fig.~\ref{fig: vis}, and additional details can be found in Appendix C.

\noindent\textbf{Results Analysis.} We then evaluate the methods on our generated nu-Caption dataset and report the result in Table \ref{tab: cap}. 
LiDAR-LLM outperforms previous 2D MLLMs for all evaluation metrics.
Specifically, our model achieves 19.26\% BLEU-4 and 91.32\% Bert score, surpassing Mini-GPT4, which achieved 2.63\% BLEU-4 and 84.38\% Bert score; also get a  11.81\% BLEU-4 and 4.87\% Bert score improvement compared to LLaMA-AdapterV2. 
These results indicate that employing 2D MLLMs directly to LiDAR data produces unsatisfactory results, leading to the omission of crucial details in caption descriptions. Concurrently, this set of results affirms that our LiDAR-LLM exhibits basic 3D scene understanding capabilities, proficiently expressing geometric relationships and engaging in common reasoning with sparse LiDAR data.
In conclusion, although 2D MLLMs achieve superior performance in the imagery domain, they are not an expert at 3D LiDAR outdoor scenes, which further validates the necessity to develop large language models specialized for LiDAR data.

\begin{table}[t]
\begin{center}
\centering
    \setlength\tabcolsep{0.05cm}
    \small
    \begin{tabular}{c|c|cc|c}
    \toprule
   Tasks& Models   & ACC-19 & ACC-5&mIoU \\
\hline
\multirow{3}{*}{3D Grounding}& Mini-GPT4~\cite{zhu2023minigpt} & 5.1   & 21.2 &-  \\
 & LLaVA1.5~\cite{liu2023improvedllava}  &9.2  &22.7&-   \\
 & Instruct-BLIP~\cite{li2023blip2}  &8.4  &23.9&-   \\
 & LLaMA-AdapterV2~\cite{zhang2023llama}  &7.1  &23.4 & -   \\
 & Ours & 34.4 & 63.1& 14.3  \\
    \bottomrule
    \end{tabular}
\end{center}
\vspace{-0.3cm}
\caption{\label{tab: grounding} 
    Experimental results on the nu-Grounding dataset. ACC-19 and ACC-5 denote the mean Top-1 accuracy for scenarios with 19 categories and 5 categories, respectively. The mIoU is calculated for the "Car" category. 
    }
\vspace{-0.2cm}
\end{table}

\begin{table*}[t]
\begin{center}
\centering
    \setlength\tabcolsep{0.15cm}
    \small
    \begin{tabular}{c|c|ccc|ccc|ccc|ccc|ccc|c}
    \toprule
\multirow{2}{*}{Method} & \multirow{2}{*}{Pretrain}  & \multicolumn{3}{c}{Exist} & \multicolumn{3}{c}{Count} & \multicolumn{3}{c}{Object} & \multicolumn{3}{c}{Status} & \multicolumn{3}{c}{Comparison} & \multirow{2}{*}{Acc} \\
& & H0 & H1 & All & H0 & H1 & All & H0 & H1 & All & H0 & H1 & All& H0 & H1 & All&   \\ 
\hline
Ex1 & None &69.4&62.3&65.5&10.4&9.7&10.0&54.8 &31.1&34.5&23.0&32.5&29.2&47.8&54.2&53.8&41.2 \\
 Ex2& C &80.0 & 71.7 & 75.5 & 13.8 & 12.6 & 13.2 & 59.6 &32.0 &36.0 &45.8 &40.8 &42.5 &73.9 &54.8 &56.2 &47.4 \\
 Ex3& G &79.7 &69.0 &73.9& 12.7 &12.0 &12.4 &58.6 &33.9 &37.4 &46.0 &34.7 &38.6 &62.6 &55.3 &55.9 &46.5 \\
 Ex4& C+G &79.1&70.6&74.5&15.3&14.7&15.0&59.6&34.1&37.8&53.4&42.0&45.9&67.0&57.0&57.8&48.6 \\
 \bottomrule
    \end{tabular}
\end{center}
\vspace{-0.4cm}
\caption{\label{tab:nus-qa} 
    The high-level instruction results on nuScenes-QA. "C" and "G" denote loading the pre-trained model parameters from the 3D captioning task and 3D grounding task, respectively. H0 and H1 represent zero-hop and one-hop reasoning questions, respectively.}
\end{table*}

\subsection{3D Grounding}
\label{sec:exp2}


\textbf{Dataset Construction.}  Apart from the captioning task, 3D grounding requires the model to have superior capabilities in object perception. Due to the absence of a grounding dataset tailored for outdoor LiDAR data, we first utilize the annotations from the nuScenes dataset to construct a comprehensive dataset, named nu-Grounding.
It consists of 280K pairs of questions and answers for both visual grounding and grounded captioning tasks, with 232K pairs allocated to the training set and 48K pairs to the validation set.
For Grounded Captioning, we evaluate accuracy separately for scenarios with 19 categories and 5 categories. For Visual Grounding, we only focus on predicting bounding boxes for the "Car" category and compute the associated BEV mIoU.

\noindent\textbf{Results Analysis.} For Grounded Captioning, as shown in Table~\ref{tab: grounding}, our model achieves 63.1\% accuracy in scenarios with 5 categories, surpassing LLaMA-Adapter and MiniGPT4 with accuracies of 23.4\% and 21.2\%, respectively. Meanwhile, when trained and tested on 19 categories, our approach still demonstrates a significant advantage over 2D MLLMs. This indicates that our LiDAR-LLM possesses an understanding of localization information in 3D scenes and achieves object classification ability under LiDAR data.
For visual grounding, our LiDAR-LLM achieves 14.3\% BEV mIoU. 
The results demonstrate that our approach not only exhibits basic perceptual capabilities but also effectively generates bounding boxes. 
More visual grounding results for other categories are shown in Appendix A.
For 3D grounding task, our goal is not solely to obtain localization information for objects but also to enhance the model's understanding of the spatial relationships within LiDAR data. This training stage further enhances our potential in high-level instruction tasks.




\subsection{High-level Instruction Task}

\label{sec:exp3}

\textbf{Dataset}
nuScenes-QA  \cite{qian2023nuscenes} is a multi-modal visual question answering benchmark for autonomous driving scenarios, which contains five types of questions: existence, counting, query-object, query-status, and comparison. The questions can be classified into two types based on reasoning complexity: zero-hop and one-hop reasoning.

\begin{figure*}[t]
\centering
\includegraphics[width=\linewidth]{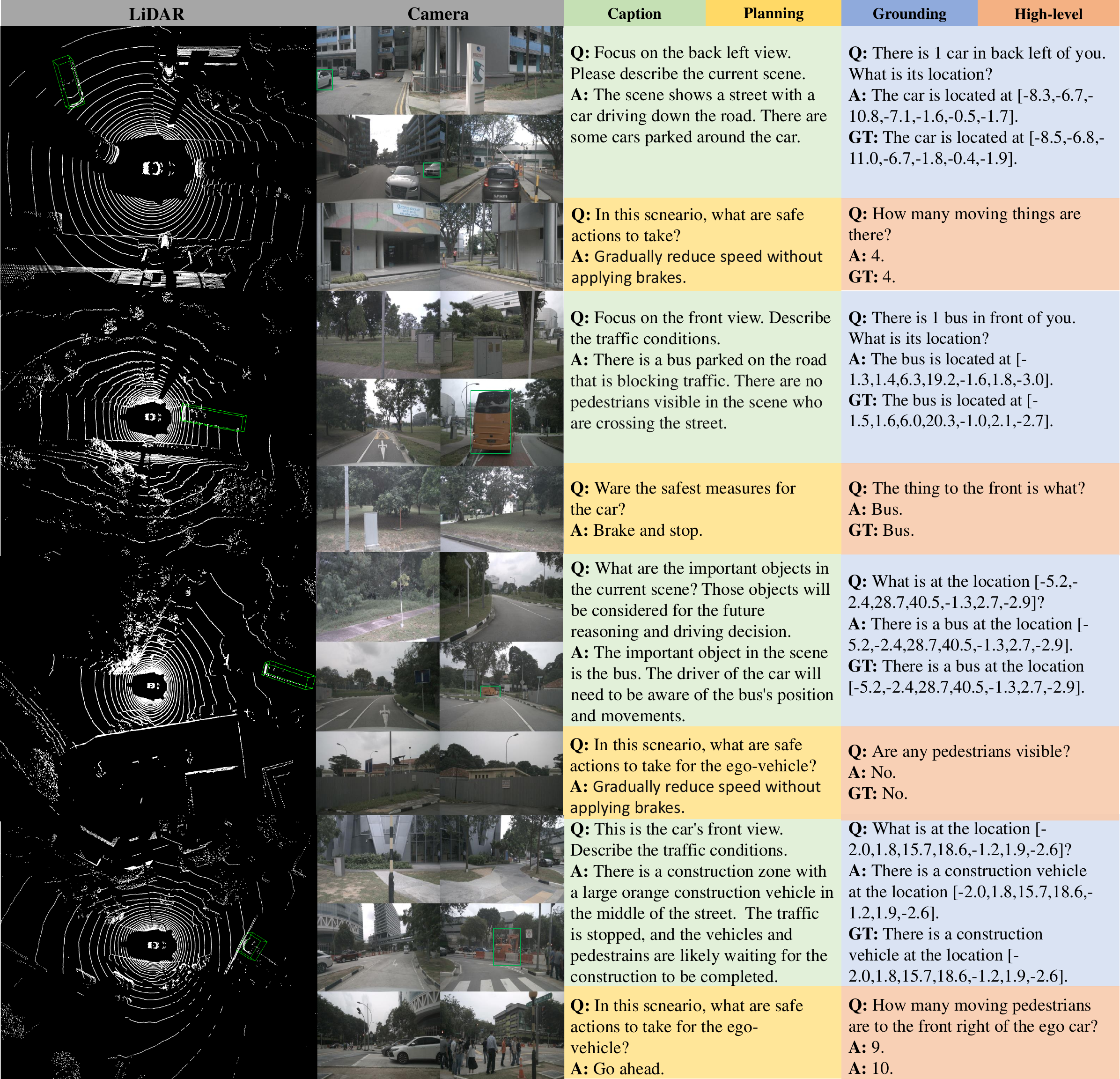}
\vspace{-0.55cm}
\caption{
Qualitative examples of prompt questions and LiDAR-LLM’s prediction. Additional examples are shown in Appendix B.} 

\label{fig: vis} 
\vspace{-0.2cm}
\end{figure*}

\textbf{Results Analysis.} 
As illustrated in Table~\ref{tab:nus-qa}, we compare our model's performance on nuScenes-QA with different pre-training stages. 
To ensure a fair comparison, we perform an equal number of fine-tuning iterations on the nuScenes-QA dataset, irrespective of which pre-trained parameters are loaded.
In Ex1, training LiDAR-LLM from scratch achieves an accuracy of 41.2\%, validating the effectiveness of our model design in high-level instruction tasks. Ex2, pre-training on the captioning task, results in a 6.2\% accuracy improvement compared to Ex1. Ex3, pre-training on the grounding task, achieves a 5.3\% accuracy improvement over Ex1. 
The results show that the first two stages of our training strategy effectively improve accuracy in the final high-level instruction task. This implies that when LiDAR-LLM possesses basic 3D scene understanding or perception capabilities, it can better accomplish high-level reasoning tasks.
Compared to pre-training on a single task, pre-training on both captioning and grounding tasks (Ex4) shows significant improvement, achieving a total accuracy of 48.6\%. This improvement is observed in both zero-hop and one-hop reasoning questions. For comparative experiments with 2D MLLMs, please refer to Appendix A.
Furthermore, we investigate the zero-shot planning capability of LiDAR-LLM. As illustrated in Figure \ref{fig: vis}, the model can generate planning-related language instructions without the need for any fine-tuning on a planning dataset.



\subsection{Ablation Study}
\label{sec:abla}

\begin{table}[!tb]
\label{ablationDAP}
\centering
\setlength\tabcolsep{4.0pt}
\renewcommand\arraystretch{1}
\begin{tabular}{c|cc|cc}
\toprule
 & \makecell*[c]{QTrans} & \makecell*[c]{VPE} &  BERT Score & BLEU-4\\\midrule
$Ex_{1}$ &  & &88.14& 11.37 \\ 
$Ex_{2}$&\checkmark  &  & 90.60& 15.41\\
$Ex_{3}$ &\checkmark    & \checkmark& 91.32& 19.26\\

\bottomrule
\end{tabular}
\vspace{-0.2cm}
\caption{Ablation study of VAT, where VPE means View Position Embedding, and QTrans means the Query Transformer. 
}
\vspace{-0.1cm}
\label{tab:ablation}
\end{table}

\textbf{Effectiveness of View-Aware Transformer (VAT).} 
To demonstrate the effectiveness of each component in VAT, we compare BLEU-4 and BERT scores in the 3D captioning task. As shown in Table.~\ref{tab:ablation}, $Ex_{1}$ denotes the absence of any transformer structure and position embedding. In this case, the BEV Feature is directly input to an MLP to align the dimensions and is then fed to the LLM, yielding only 88.14\% BERT score and 11.37\% BLEU-4. In $Ex_2$, with the use of Query Transformer, we observe that the BERT Score achieves 90.60\% and 15.41\% BLEU-4. 
The results demonstrate that the structure of the Query Transformer can better align LiDAR features with text embeddings. Compared with $Ex_2$ and $Ex_3$, the introduction of the View Position embedding achieves an 0.7\% BERT score and 3.85\% BLEU-4 improvement. This set of results affirms that incorporating the View Position embedding aids the model in better understanding 3D scenes and spatial relationships.

\textbf{Effectiveness of the three-stage training strategy.} 
As shown in Table~\ref{tab:nus-qa}, we validate the effectiveness of our proposed three-stage training strategy on the high-level instruction task (NusceneQA). We decompose the strategy into captioning, grounding, and the final high-level task. For detailed experimental analysis, please refer to Section \ref{sec:exp3}.

\subsection{Qualitative Analysis}
\label{sec:vis}
In the 3D captioning task, as indicated in green in Fig.~\ref{fig: vis}, LiDAR-LLM demonstrates its proficiency in aligning linguistic and LiDAR input, showcasing an understanding of contextual information within the LiDAR data, and providing answers based on both textual questions and corresponding visual information.
From the figure, our method excels in identifying crucial objects and evaluating their states, such as ``parked" or ``driving down," showcasing its ability in comprehending 3D scenes.
During the grounding stage, as indicated in blue in Fig.~\ref{fig: vis}, the model showcases its ability to identify the referred object, exhibit spatial relation awareness, and demonstrate precise localization skills.
In the yellow section of Fig.~\ref{fig: vis}, we illustrate the interpretability of LiDAR-LLM in planning tasks.
Based on the capabilities unleashed by our method, LiDAR-LLM generates a coherent high-level action. For instance, in the second sub-figure, we prompt the model to describe the safest measures for the car. The model precisely identifies the critical object, the 'bus,' and analyzes its potential impacts on the ego car. With these reasoning priors, when instructing the model to plan a 'safest measures', it generates 'brake and stop' to avoid a collision with the obstructing bus.
Furthermore, in another high-level task (NuScenes-QA \cite{qian2023nuscenes}), as represented by the pink section in Fig.~\ref{fig: vis}, LiDAR-LLM demonstrates strong capabilities by accurately counting objects, identifying object classifications, and reasoning about spatial relationships among surrounding objects.


\section{Conclusion}
\label{sec: con}

In conclusion, our paper represents a pioneering effort to unleash the reasoning capabilities of LLMs to comprehend outdoor LiDAR data. Our proposed LiDAR-LLM reformulates the intricate challenge of 3D outdoor scene understanding as a language modeling problem. 
To train LiDAR-LLM, we generate a comprehensive set of LiDAR-text paired datasets, encompassing 420K 3D captioning and 280K 3D grounding data. 
We then introduce a three-stage training strategy, involving cross-modal alignment, perception, and high-level instruction, aligning the LiDAR modality with the language embedding space of the LLM.
Our architectural innovation introduces the View-Aware Transformer (VAT) to connect the 3D encoder with the LLM. This design effectively bridges the modality gap and enhances the LLM's comprehension of spatial orientation in LiDAR features.
Through extensive experimentation on both our generated datasets and open-source datasets, our LiDAR-LLM demonstrates promising performance across diverse tasks, including 3D captioning, 3D grounding, 3D question answering, and autonomous driving planning.
In future work, we will explore the continual transfer learning \cite{yang2023exploring, gan2023cloud, liu2023vida} and lightweight operations \cite{li2023unlock, chang2023detrdistill} of MLLMs, making it feasible to deploy MLLMs on edge devices.

{
    \small
    \bibliographystyle{ieeenat_fullname}
    \bibliography{main}
}

\clearpage

\appendix

\begin{table*}[t]
\begin{center}
\centering
    \setlength\tabcolsep{0.25cm}
    \small
    \begin{tabular}{c|c|cccccc}
    \toprule
    Tasks& Models   & BLEU-1& BLEU-2& BLEU-3& BLEU-4 & Bert Score \\
\hline
\multirow{5}{*}{3D Captioning}& Mini-GPT4 & 14.97   & 6.76   & 3.74 & 2.63 &84.38\\
 & LLaVA  & 19.92  &12.10 &8.57&5.37  &85.01 \\
 & Instruct-BLIP  & 18.67 &13.38 &7.41&5.20  &85.89 \\
 & LLaMA-AdapterV2  & 30.17  &17.34 &10.40&7.45  &86.45 \\
 
 & Ours & 40.98& 29.96 & 23.43 & 19.26   &91.32\\
    \bottomrule
    \end{tabular}
\end{center}
\vspace{-0.5cm}
\caption{\label{tab:sup_caption} 
    Experimental Results on nu-Caption dataset. Our model outperforms all baseline models for all evaluation metrics.}
\end{table*}

\textbf{Supplementary Materials - Overall}

Due to space constraints in the main paper, we provide a thorough quantitative and qualitative analysis of the proposed method in this supplementary material.
In Appendix~\ref{sec: supa}, we offer more extensive experiments in 3D captioning, 3D grounding, and the High-level Instruction task. In Appendix~\ref{sec: supb}, additional qualitative analyses are presented across multiple downstream tasks, encompassing both good and challenging cases in question answering. Finally, in Appendix~\ref{sec: supc}, we showcase the details of our generated nu-Caption and nu-Grounding datasets.

\section{Additional Experiments}
\label{sec: supa}
In this section, our aim is to supplement additional baseline comparison experiments, including Mini-GPT4 \cite{zhu2023minigpt}, LLaVA \cite{liu2023improvedllava}, Instruct-BLIP \cite{li2023blip2}, and LLaMA-AdapterV2 \cite{zhang2023llama}. Simultaneously, in Section \ref{sec: sup3DG}, we extend the visual grounding experiment to include a broader range of categories.

\subsection{3D Captioning}

As shown in Table~\ref{tab:sup_caption}, we evaluate the methods on our generated nu-Caption dataset. Specifically, compared to the previous 2D MLLMs model, our model achieves 19.26\% BLEU-4 and 91.32\% Bert score, surpassing the previous sota method LLaMA-AdapteV2 an 11.81\% BLEU-4 and 4.87\% Bert score improvement. 

These results suggest that the direct application of 2D MLLMs to LiDAR data yields unsatisfactory outcomes, resulting in the exclusion of vital information in caption descriptions. Concurrently, this set of findings confirms that our LiDAR-LLM demonstrates fundamental 3D scene understanding abilities, effectively capturing geometric relationships and engaging in common reasoning with sparse LiDAR data.

\subsection{3D Grounding}
\label{sec: sup3DG}
\begin{table}[t]
\begin{center}
\centering
    \setlength\tabcolsep{0.07cm}
    \small
    \begin{tabular}{c|c|cc}
    \toprule
   Tasks& Models   & ACC-19 & ACC-5 \\
\hline
\multirow{5}{*}{3D Grounding}& Mini-GPT4 & 5.1   & 21.2  \\
 & LLaVA  &9.2  &22.7   \\
 & Instruct-BLIP  &8.4  &23.9   \\
 & LLaMA-AdapterV2  &7.1  &23.4  \\
 & Ours & 34.4 & 63.1  \\
    \bottomrule
    \end{tabular}
\end{center}
\vspace{-0.5cm}
\caption{\label{tab: sup_grounding} 
    Experimental results on the nu-Grounding dataset. ACC-19 and ACC-5 denote the mean Top-1 accuracy for scenarios with 19 categories and 5 categories, respectively. 
    }
\end{table}

\begin{table}[t]
\begin{center}
\centering
    \setlength\tabcolsep{0.08cm}
    \small
    \begin{tabular}{c|ccccc}
    \toprule
   & Car  & Pedestrian & Bus & Truck & Construction\_vehicle \\
\hline
mIoU & 11.94 & 9.05 & 11.23& 8.09& 9.40\\
    \bottomrule
    \end{tabular}
\end{center}
\vspace{-0.5cm}
\caption{\label{tab: sup_box} 
    The visual grounding results.
    }
\end{table}

\begin{table*}[t]
\begin{center}
\centering
    \setlength\tabcolsep{0.15cm}
    \small
    \begin{tabular}{c|ccc|ccc|ccc|ccc|ccc|c}
    \toprule
\multirow{2}{*}{Method}  & \multicolumn{3}{c}{Exist} & \multicolumn{3}{c}{Count} & \multicolumn{3}{c}{Object} & \multicolumn{3}{c}{Status} & \multicolumn{3}{c}{Comparison} & \multirow{2}{*}{Acc} \\
 & H0 & H1 & All & H0 & H1 & All & H0 & H1 & All & H0 & H1 & All& H0 & H1 & All&   \\ 
\hline
 LLaMA-AdapterV2 & 34.2&6.3&19.3&5.0&0.1&2.7&23.7&4.6&7.6&9.8&11.3&10.8&2.6&1.5&1.6&9.6 \\
 LLaVA&38.9& 51.9 & 45.8 &7.7 &7.6 &7.7 &10.5 &7.4 &7.8 &7.0 &9.9 &9.0 &64.5 &50.8 &52.1&26.2 \\
Ours&79.1&70.6&74.5&15.3&14.7&15.0&59.6&34.1&37.8&53.4&42.0&45.9&67.0&57.0&57.8&48.6 \\
 \bottomrule
    \end{tabular}
\end{center}
\vspace{-0.6cm}
\caption{\label{suptab:nus-qa} 
    The high-level instruction results on nuScenes-QA. H0 and H1 represent zero-hop and one-hop reasoning questions, respectively.}
\end{table*}

\begin{figure*}[t]
\centering
\includegraphics[width=\linewidth]{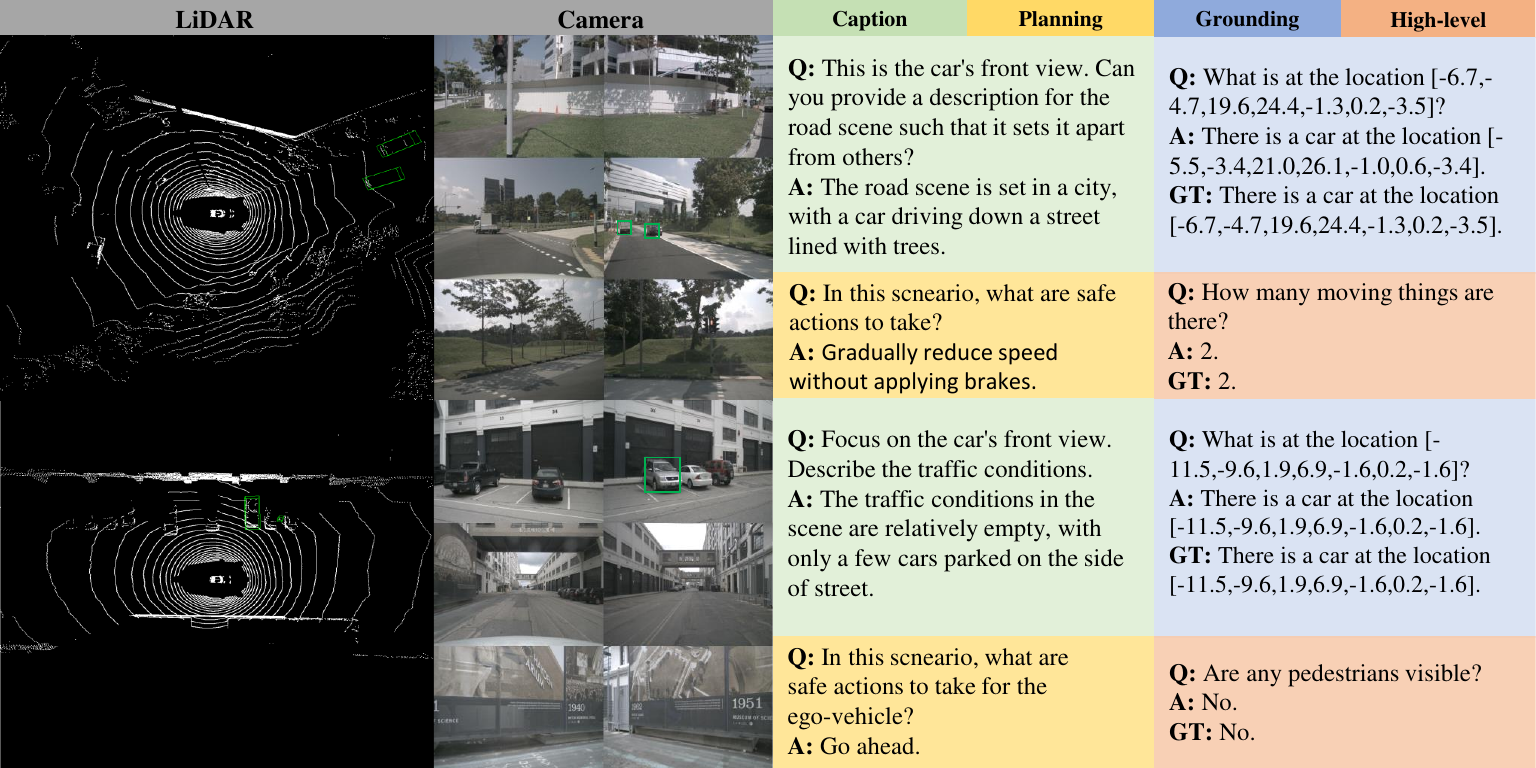}
\vspace{-0.55cm}
\caption{
Some correct qualitative examples of LiDAR-LLM’s predictions.} 
\label{fig: good} 
\vspace{-0.2cm}
\end{figure*}

\begin{figure*}[t]
\centering
\includegraphics[width=\linewidth]{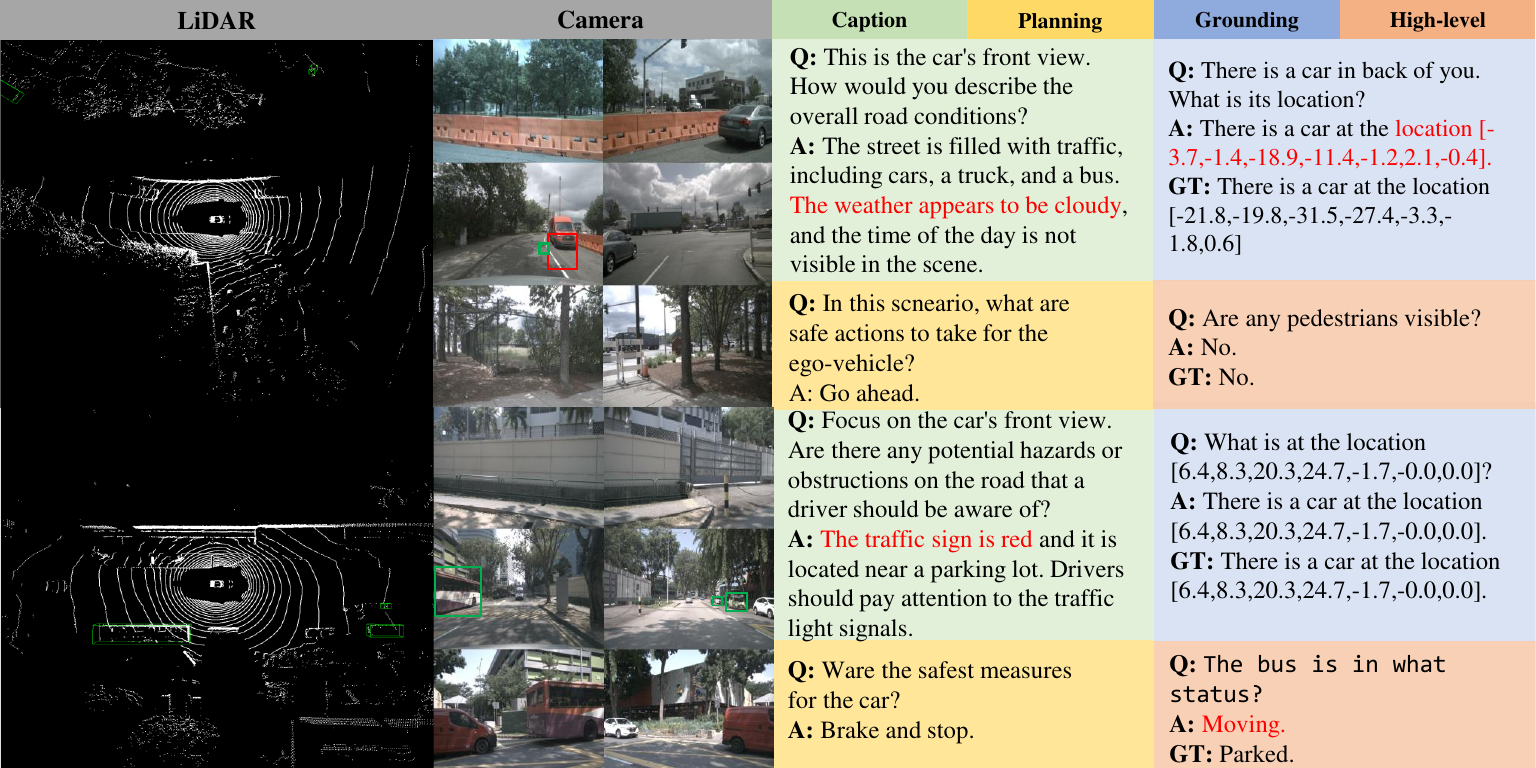}
\vspace{-0.55cm}
\caption{
Some failure qualitative examples of LiDAR-LLM’s predictions.} 
\label{fig: bad} 
\vspace{-0.2cm}
\end{figure*}

\begin{table*}[h]
\begin{tcolorbox}[colback=gray!10,
                  colframe=black,
                  width=\textwidth,
                  arc=1mm, auto outer arc,
                  boxrule=0.5pt,
                 ]

\textcolor{blue}{\textbf{The general description of current scenes:}}

\textcolor{red}{Question:} ["This is the car's back right view. How would you uniquely describe the road scene?"]

\textcolor{red}{Answer:} ["The road scene features a one-way street with a car driving on it, a motorcycle parked on the side, and a person riding a bicycle. The presence of a car, motorcycle, and bicycle on the same street suggests that the area is a mix of different modes of transportation, which is not common in many urban settings."]

\textcolor{blue}{\textbf{The detailed description of objects:}}

\textcolor{red}{Question:} ["This is the car's back left view. In the view, are there any moving objects, and provide a description of them."]

\textcolor{red}{Answer:} ["Yes, there are cars moving in the background of the scene. They are likely driving on the road, and their presence adds to the overall activity and movement in the scene."]

\textcolor{blue}{\textbf{The recognition of potential risks:}}

\textcolor{red}{Question:} ["This is the car's back left view. From the current scene, identify any potential challenges a driver might face."]

\textcolor{red}{Answer:} ["In this scene, a car is parked on a street next to a forest. The driver might face challenges such as limited visibility due to the trees and the dense forest. The presence of trees and branches on the street could obstruct the driver's view of other vehicles, pedestrians, or obstacles, increasing the risk of accidents."]
\end{tcolorbox}
\caption{Prompt examples of 3D LiDAR Captioning}
\label{tab: EX3dc}
\end{table*}

\begin{table*}[ht]
\begin{tcolorbox}[colback=gray!10,
                  colframe=black,
                  width=\textwidth,
                  arc=1mm, auto outer arc,
                  boxrule=0.5pt,
                 ]

\textcolor{blue}{\textbf{Grounded Captioning:}}

\textcolor{red}{Question:} ["What is at the location [-9.0,-8.7,27.6,27.9,-1.0,-0.2,-1.2]?"]

\textcolor{red}{Answer:} ["There is a traffic cone at the location [-9.0,-8.7,27.6,27.9,-1.0,-0.2,-1.2]."]

\textcolor{red}{Question:} ["What is at the location [-14.7,-11.5,30.3,43.7,-0.6,3.0,-3.5]?"]

\textcolor{red}{Answer:} ["There is a construction vehicle at the location [-14.7,-11.5,30.3,43.7,-0.6,3.0,-3.5]."]

\textcolor{red}{Question:} ["What is at the location [-14.7,-11.5,30.3,43.7,-0.6,3.0,-3.5]?"]

\textcolor{red}{Answer:} ["There is a construction vehicle at the location [-14.7,-11.5,30.3,43.7,-0.6,3.0,-3.5]."]

\textcolor{blue}{\textbf{Visual Grounding:}}

\textcolor{red}{Question:} ["There is 1 pedestrian in front of you. What is its location?"]

\textcolor{red}{Answer:} ["The pedestrian is located at [[26.6,27.3,-11.7,-11.1,-0.5,1.3,-1.7]]."]

\textcolor{red}{Question:} ["There are 2 cars in front left of you. What are their locations?"]

\textcolor{red}{Answer:} ["The 2 cars are located at [[1.0,2.8,-31.3,-26.7,-2.3,-0.8,-3.3],[3.7,5.5,-6.5,-1.9,-1.9,-0.4,0.0]]."]
\end{tcolorbox}
\caption{Prompt examples of 3D LiDAR Grounding}
\label{tab: Grounding}
\end{table*}

 For Grounded Captioning, as shown in Table~\ref{tab: sup_grounding}, our model achieves 63.1\% accuracy in scenarios with 5 categories, significantly surpassing the previous sota method Instruct-BLIP with accuracies of 23.9\%. Meanwhile, to further demonstrate the effectiveness of our LiDAR-LLM, we trained and tested on 19 categories, our approach still showcases a significant advantage over 2D MLLMs, which have more than 25\% improvement. This indicates that our LiDAR-LLM possesses an understanding of localization information in 3D scenes and achieves object classification ability under LiDAR data.

As shown in Table~\ref{tab: sup_box}, we extend the visual grounding task to include location outputs for five categories, namely car, pedestrian, bus, truck, and construction vehicle.
The results indicate that our method efficiently produces object localization based on the question prompt. In comparison to the single category presented in the main paper, our method does not decrease the BEV mIoU even with the expansion to five categories. Specifically, for the challenging pedestrian category, our method achieves a mIoU of 9.05\%.
We refrain from using the LLM tokenizer to directly generate the 3D bounding box, as it can only leverage the classification loss (cross entropy) to optimize the word token corresponding to the box location. To attain a more accurate bounding box result, we project the last hidden state feature of the LLM into an MLP network and employ Regression Loss (L2 loss) for optimization.

\subsection{High-level Instruction Task}
As illustrated in Table~\ref{suptab:nus-qa}, we compare our model's performance on nuScenes-QA with previous 2D sota MLLMs models.  The results highlight a significant improvement in our model compared to the previous method. Notably, our model achieves an accuracy of 45.9\% for the status task, showcasing a substantial advantage over the previous SOTA method which only gained 9.0\% accuracy.
These results show the capability of our model to effectively handle high-level reasoning tasks.

\section{Additional Qualitative Analysis} 
\label{sec: supb}
We have included additional samples for visualization, as shown in Fig.~\ref{fig: good} and Fig.~\ref{fig: bad}.
\textbf{(a) Good cases.}
In Fig.~\ref{fig: good}, we show LiDAR-LLM's proficiency in aligning linguistic and LiDAR input, and comprehend the LiDAR scene. 
For example, given the top scene in Fig.~\ref{fig: good}, under the captioning task (green section), the model accurately interprets the scene as a ``street lined with trees" and deduces that it is ``set in a city''. 
It demonstrates the ability to extract information from the LiDAR data, respond to textual queries, and reason about its understanding of the scene. 
Moreover, LiDAR-LLM consistently generates coherent high-level actions. 
For example, in the bottom scene of Fig.~\ref{fig: good}, when tasked with providing the safest measures for the car, the model correctly identifies the ``traffic conditions are empty with only a few cars parked on the side" during the captioning task. 
This indicates its observation and assessment of potential risks to the ego car. Therefore, guided by these reasoning priors, when planning ``safe measures," it generates the action ``go ahead," as such action avoids unsafety risks.
\textbf{(b) Bad cases.}
Inevitably, there are also instances of failure, as illustrated in Fig.~\ref{fig: bad}. Due to the inherent limitation of LiDAR in providing rich semantic information compared to cameras, LiDAR-LLM may exhibit ``illusions" when confronted with elements beyond the scope of the LiDAR coverage. For instance, in the captioning task for the top scene, LiDAR-LLM erroneously describes a clear and sunny scene as ``cloudy" due to the absence of color information. Additionally, in the bottom scene, LiDAR-LLM predicts a ``red" traffic light that does not even exist, as LiDAR data lacks color information for the model to learn from. Furthermore, there are errors in grounding and high-level action tasks, which are highlighted in red.

\section{LiDAR-Language Data} 
\label{sec: supc}
In this section, we will showcase the details of our generated nu-Caption datasets and nu-Grounding datasets.

\textbf{nu-Caption}  As shown in Table~\ref{tab: EX3dc}, the nu-Caption follows a progression of increasing complexity. Initially, it offers a broad depiction of current scenes or traffic conditions. Subsequently, it delves into finer details, providing a comprehensive and intricate description of objects and their relationships. Finally, it demonstrates its ability to recognize potential risks on the road.

To enhance the quality of the questions aligned with these captions, we utilize a systematic approach by leveraging the LLaMA-Adapter and GPT-4.
We begin by structuring our questions in humans. Subsequently, we employ the capabilities of GPT-4 to refine and extend these questions, ensuring that they meet the elevated standard of quality required.
Additionally, to generate high-quality QA-pairs, we initially utilize the LLaMA-AdapterV2 to generate the original caption. We then employ GPT-4 to refine and enhance the QA pairs. 
Consequently, we have collected a total of 349K LiDAR-text pairs for the training set and 72K pairs for the validation set. 

\textbf{nu-Grounding} To further endow the model with instance-level perception abilities, we collect the nu-Grounding dataset.
The nu-Grounding dataset consists of two main aspects: Grounded Captioning and Visual Grounding.
As shown in Tabel~\ref{tab: Grounding}, in grounded captioning, the questions comprise bounding boxes and aim to generate captions that specifically describe localized areas. On the other hand, Visual Grounding focuses on raising objects in the questions and requires the model to provide bounding box answers. These challenging datasets, characterized by bounding boxes with 7 data points [xmin, ymin, zmin, xmax, ymax, zmax, angle], serve to enhance the model's instance-level perception. All the nu-Grounding data are collected through the Ground-truth of the nuScenes dataset.

\end{document}